\documentclass[runningheads]{llncs}
\input{psfig.sty}
\usepackage{algorithm,algpseudocode}
\usepackage{graphicx}
\usepackage{url}
\usepackage{xcolor}
\usepackage{etoolbox}
\patchcmd{\thebibliography}{\chapter*}{\section*}{}{}
\graphicspath{ {images/} }

\emergencystretch=\maxdimen
\begin{document}

\pagestyle{headings}

\mainmatter

 \title{Input Fast-Forwarding\\
 for Better Deep Learning}
 \titlerunning{Lecture Notes in Computer Science}
 \author{Ahmed Ibrahim\inst{1,4}, A. Lynn Abbott\inst{1}, Mohamed E. Hussein\inst{2,3} \\
 \institute{Virginia Polytechnic Institute and State University, USA\\
 \and Egypt-Japan University of Science and Technology, Egypt\\
 \and Alexandria University, Egypt\\
 \and Benha University, Egypt \\
\email{\{nady,abbott\}@vt.edu; mohamed.e.hussein@ejust.edu.eg}}
 }

\maketitle
\begin{abstract} 
    This paper introduces a new  architectural framework, known as input fast-forwarding, that can enhance the performance of deep networks. The main idea is to incorporate a parallel path that sends  representations of  input values forward to deeper network layers. This scheme is substantially different from ``deep supervision,'' 
in which the loss layer is re-introduced to  earlier layers. 
The parallel path provided by fast-forwarding enhances the training process in two ways. First,  it enables the individual layers to combine  higher-level information (from the standard processing path) with lower-level  information (from the fast-forward path).
Second, this new architecture reduces the problem of vanishing gradients substantially because 
the fast-forwarding path provides a shorter route for  gradient backpropagation.
In order to evaluate the utility of the proposed technique, a Fast-Forward Network (FFNet), with 20 convolutional layers along with parallel fast-forward paths, has been created and tested. 
The paper  presents empirical results that demonstrate improved learning capacity of FFNet due to fast-forwarding, as compared to
GoogLeNet (with deep supervision) and CaffeNet, which are $4\times$ and $18\times$ larger in size, respectively.
 All of the source code and deep learning models described in this paper will be made  available to the entire research community\footnote{https://github.com/aicentral/FFNet}.
\end{abstract}

\section{Introduction}
Developments in deep learning have led to networks that have grown from
5 layers in LeNet \cite{lenet}, introduced in 1998, to 152 layers in the latest version of ResNet \cite{resnet}. 
One consequence of deeper and deeper  networks is the problem of vanishing gradients during training. 
This problem occurs as error values, which depend on the computed gradient values, are  propagated backward through the network to update the weights at each layer.
With each additional layer, a smaller fraction of the error gradient is available to guide the adjustment of network weights.
As a result, the weights in early layers are updated very slowly; hence, the performance of the entire training process is degraded.

Many models have been proposed to overcome the vanishing-gradient problem.
One approach is to provide  alternative paths for signals to travel, as compared to traditional layer-to-layer pathways.
An example of this approach is the Deeply-Supervised Network (DSN) \cite{dsn}, where a companion objective function is added to each hidden layer in the network, providing gradient values directly to the hidden layers. DSN uses Support Vector Machines (SVM) \cite{SVM} in its companion objective function, which means that  end-to-end training of the network is not supported. Another example is relaxed deep supervision \cite{relaxedlabels}, where an improvement over a holistic edge detection model \cite{hed} is made by providing relaxed versions of the target edge map to the earlier layers of the network. This approach provides  a version of the gradient directly to the early layers. However, relaxed deep supervision is suitable only for problems where relaxed versions of the labels can be created, such as for maps of intensity edges. GoogLeNet \cite{googlenet} is another model that uses a mechanism to address the problem of vanishing gradients. More relevant details about GoogLeNet will be given in section \ref{section:relatedwork} because it serves as a baseline for comparison with our proposed model.

The novel approach that is proposed here provides parallel signal paths that carry 
simple representations of the input to deeper layers through what we call a fast-forwarding branch. This approach allows for a novel integration of ``shallower information'' with ``deeper information'' by the network.
During training the fast-forwarding branch provides an effective means for back-propagating errors so that the vanishing-gradient problem is reduced.

To demonstrate the efficacy of the model, we created a 20 layer network with fast-forwarding branches, which we call FFNet. To study the effect of the fast-forwarding concept, the network layers are made of simple convolutional layers followed by fully connected layers with no additional complexities. 
The results that we have obtained using the 
the relatively small and simple FFNet model have been surprisingly good, especially when compared with the performance of bigger and more complex models. 

The rest of this paper is organized as follows. 
Section \ref{section:relatedwork} presents a brief survey of related work, including a discussion of the models that will be used as a baseline to be compared with FFNet.
Section \ref{proposedmodel} provides details concerning the proposed model.
In order to gauge the performance of this approach, experimental results from FFNet were compared with results from several well-known deep models. These experiments are described in Section \ref{eval}. Finally, concluding remarks are given in Section \ref{section:conclusion}.

\section{Related Work}
\label{section:relatedwork}

\subsection{Deep Learning}
Deep learning is a  machine-learning technique that has become increasingly popular in computer vision research. 
The main difference between classical machine learning (ML) and deep learning is the way that features are extracted. 
For classical ML techniques such as support vector machines (SVM)~\cite{SVM}, feature extraction is performed in advance using techniques crafted by the researchers. Then, the training procedure develops weights or rules that map any given feature vector to an output class label. 
In contrast, the typical deep-learning procedure is to directly feed signal values as inputs to the training procedure, without any preliminary efforts at feature extraction.
The network takes the input signal (pixel values, in our case), and assigns a class label based on those signal values directly. 
Because 
the deep-learning approach implicitly must derive its own features, many more training samples are required than for traditional ML approaches.

Several deep-learning packages are available for researchers. 
The popular package that we have used to evaluate the proposed model is Caffe \cite{caffe}, which was created with computer vision tasks in mind. Caffe is relatively easy to use, flexible, and powerful. It was developed in C++ using GPU optimization libraries, such as CuDNN \cite{cudnn}, BLAS \cite{openblas}, and ATLAS \cite{atlas}. In the next sections, we will discuss briefly two well-known deep models,  AlexNet and GoogLeNet. These two models will be used as a baseline for comparison with the proposed FFNet model.

\subsection{AlexNet and CaffeNet}
   AlexNet \cite{alexnet} was the first deep model to win the ILSVRC \cite{ilsvrc} challenge. For  the ILSVRC-2012 competition, AlexNet won with a top-5 test error rate of 15.3\%, compared to 26.2\% achieved by the second-best entry. This model consists of five convolutional layers followed by three fully-connected layers. The creators of Caffe \cite{caffe} introduced a slightly modified version of AlexNet by switching the order of pooling and normalization layers. They named the modified version CaffeNet \cite{caffenet}. As the only modification done to the network is switching the order of pooling and normalization layers, the size of the network is exactly the same as AlexNet.
   
   AlexNet and CaffeNet will be used to provide  baseline cases of simple architectures that rely on huge numbers of parameters.
The number of filters in the convolutional layers range from 96 to 384 in AlexNet, while the proposed FFNet model uses only 64 filters in each convolutional layer. AlexNet uses a 4069-node fully-connected layer followed by another layer of the same size, whereas FFNet uses only a 400-node fully connected layer followed by a 100-node layer. The total size of AlexNet is therefore approximately 18 times bigger than FFNet.
   
\subsection{GoogleNet}
GoogLeNet \cite{googlenet} is another winner of the ILSVRC challenge. This model won the ILSVRC-2014 competition with a top-5 test error rate of 6.6\%.  The network consists of 22 layers with a relatively complex design called ``inception.'' The inception module, which is used to implement the layers of GoogLeNet, consists of parallel paths of convolutional layers of different sizes concatenated together. The number of filters in the convolutional layers inside the inception modules ranges from 16 to 384. (By comparison, in FFNet the number of filters in each convolutional layer is fixed.) 
In addition to using the inception design, GoogLeNet uses three auxiliary classifiers connected to the intermediate layers during training. GoogLeNet is of interest to us as a baseline for comparison because of its depth, because of its complex architecture, and especially because of the auxiliary classifiers.
GoogLeNet is 4 times bigger and far more complex than the proposed FFNet model.

\subsection{Benchmarking Datasets}
Many datasets have been created to aid in machine learning for computer vision.
To evaluate the proposed FFNet model, we selected two publicly available datasets, COCO-Text-Patch and CIFAR-10.

COCO-Text-Patch \cite{cocotextpatch}, contains approximately $354,000$ images of size $32\times32$  that are each labeled as ``text'' or ``non-text.'' This dataset was created to address the problem of text verification, which is an essential stage in the end-to-end text detection and recognition pipeline. The dataset is derived from COCO-Text \cite{cocotext}, which contains $63,686$ images of real-world scenes with  $173,589$ instances of text. 

CIFAR-10 \cite{cifar10} is a labeled subset of the ``$80$ million tiny images'' dataset \cite{tiny8million}. They were collected by 
the creator of AlexNet. The CIFAR-10 dataset consists of $60,000$ color images of size $32\times32$ in $10$ classes, with $6,000$ images per class.

\section{Proposed Model: FFNet} \label{proposedmodel}
\begin{figure}
\centering
\begin{minipage}{.45\textwidth}
  \centering
\includegraphics[width=1.15\textwidth,keepaspectratio]{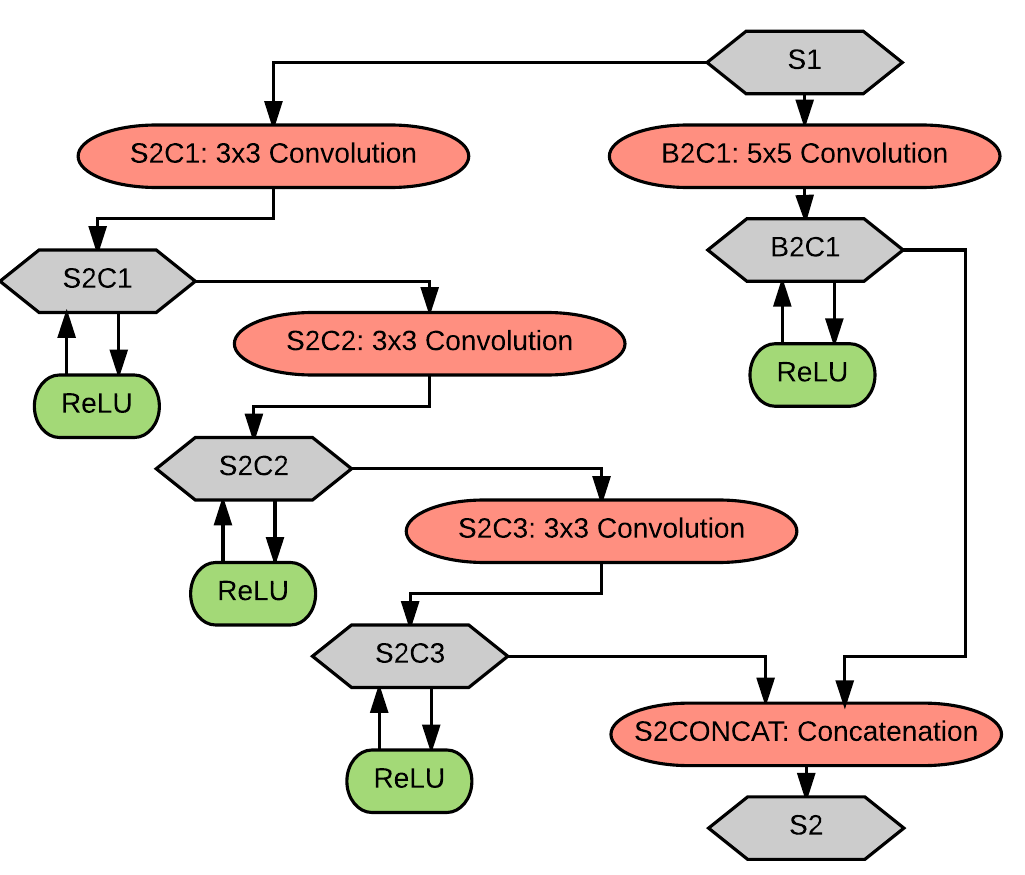}
\caption{A single fast-forwarding stage. Node $S1$ represents the input, and $S2$ is the output. The left pathway contains common convolutational blocks. At the right is the fast-forward path.}
  \label{fig:ffstage}
\end{minipage}%
\hspace{.1\textwidth}
\begin{minipage}{.44\textwidth}
  \centering
\includegraphics[width=0.5\textwidth,keepaspectratio]{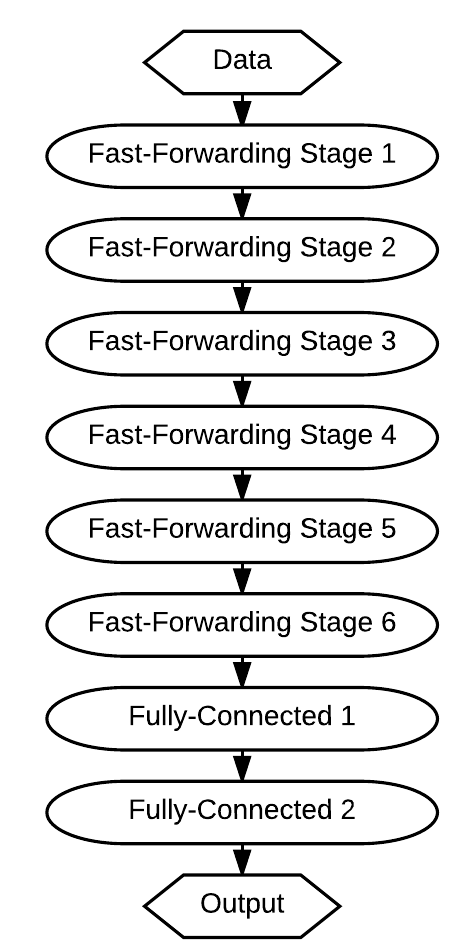}
\caption{Proposed FFNet model. Because of fast-forwarding, this relatively small network has yielded empirical results that are better than much larger deep networks.}
  \label{fig:proposedmodel}
\end{minipage}
\end{figure}

The new FFNet model consists of convolutional units that are organized into a sequence of stages. 
Within each stage, as illustrated in figure \ref{fig:ffstage}, computations are performed in 2 parallel paths.
The left branch in the figure represents a standard convolutional path, whereas the right branch represents an extra parallel data path.
It is this parallel, ``fast-forwarding'', path that delivers the improved performance of the network.

The input to the stage, $S1$, arrives from the previous layer, and the output to the next layer is shown as $S2$.
The standard (deep) branch consists of three consecutive $3\times3\times64$ convolutional layers. Each layer is followed by an in-place Rectified Linear Unit (ReLU).
The last layer of the deep branch is padded with zeros, for reasons that are described below.

Let  the input $S1$ be of size ${N}\times{N}\times{C}$. The value of $C$ is the number of channels, which is typically $128$ except for the first stage where $C=3$ to match the input data.
Refer to a stage's deep convolutional layers as $S2C1$, $S2C2$, and $S2C3$, as shown in the figure.
The deep branch's output $S2C3$ can be represented as follows,
where $CONV$ is the convolutional operation, $s$ is the stride, and $p$ is the padding: 
\begin{equation}
  S2C3 = {CONV}_{3\times3,s=1,p=1}({CONV}_{3\times3,s=1,p=0}({CONV}_{3\times3,s=1,p=0}(S1)))
\label{eq1}
\end{equation}

\noindent
The size of $S2C3$ will be $({N-2})\times({N-2})$. 

The fast-forwarding branch consists of a single $5\times5\times64$ convolutional layer followed by a ReLU. This branch takes  $S1$ as input, and generates the output $B2C1$ that can be represented as follows:
\begin{equation}
  B2C1 = {CONV}_{5\times5,s=1,p=0}(S1)
\label{eq2}
\end{equation}
\noindent
No padding is used for the fast-forwarding branch, so that the resulting output size is also $({N-2})\times({N-2})$. This branch will provide a ``shallower'' 
representation of the input $S1$ to the next stage. 

The outputs of the deep branch and of the fast-forwarding branch are concatenated to create the single stage output $S2$. The size of $S2$ will be $({N-2})\times({N-2})\times128$.
Because the last layer of the deep branch is padded with zeros, both branches provide data of the same size to the output.

To evaluate the fast-forwarding concept,
we built a Fast-Forwarding Network (FFNet) that consists of 6 consecutive fast-forwarding stages followed by two fully connected layers plus an output layer, as shown in figure \ref{fig:proposedmodel}. 
The 6 fast-forwarding stages consist of a  total of 18  convolutional layers, each of size $3\times3\times64$. The first layer of the two fully-connected layers consists of 400 nodes, while the second layer consists of 100 nodes.

\section{Evaluation}
\label{eval}
To evaluate the performance of the proposed model, a number of experiments were conducted that compare  FFNet to AlexNet, CaffeNet, and GoogLeNet. 
The publicly available datasets CIFAR-10 \cite{cifar10} and COCO-Text-Patch \cite{cocotextpatch} were used in the evaluation, as described previously.
FFNet was implemented using 
Caffe \cite{caffe}.
Standard 10-crop augmentation was applied to the datasets. 
All the training and testing were performed on a GPU with batch size 32. The training was stopped after $150,000$ iterations as the validation accuracy and loss started to plateau.

\begin{figure}[!]
\centering
\includegraphics[width=0.99\textwidth,keepaspectratio]{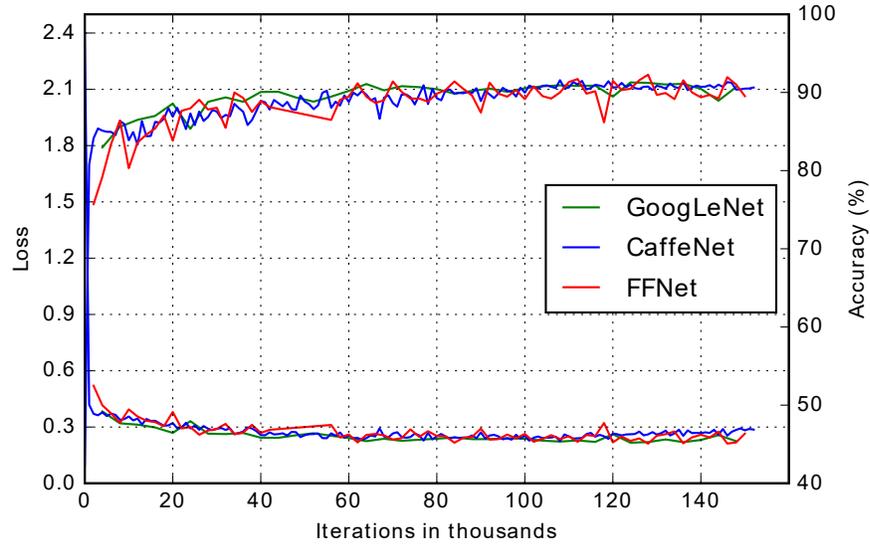}
\caption{COCO-Text-Patch validation accuracy and loss for the proposed FFNet model (red), CaffeNet (blue), and GoogLeNet (green).}
\label{fig:trainingloss}
\end{figure}

A summary of results is provided in table \ref{table_results}. Despite its relatively small size, the performance of the proposed FFNet model exceeded the performance of CaffeNet and GoogleNet in these experiments.
The accuracy and validation loss graphs shown in figure \ref{fig:trainingloss} demonstrate how the proposed model converges with the same speed as CaffeNet and GoogLeNet.
These trends provide evidence  of the effectiveness of the fast-forwarding approach in fighting the vanishing-gradient problem.

\begin{table}[]
\centering
\caption{Performance comparison of the proposed FFNet model with several common alternatives. Although FFNet is much smaller than the other models, its error rate was lower than the others (with one exception), using publicly available test sets.}
\label{table_results}
\begin{tabular}{|c|c|c|c|c|c|}
\hline
\multicolumn{4}{|c|}{\textit{Model}}  & \multicolumn{2}{c|}{\textit{Error Rate} (\%)}   \\ \hline
Description &  Layers & Size (MB) & Time*(ms) & {CIFAR-10} & {CTP**} \\ \hline
{AlexNet with dropout \cite{alexnet}}& {8}& {181.3} & - &{15.6} & -\\ \hline
{AlexNet with stoch. pooling \cite{alexstochastic}}& {8} & {181.3} & - &{15.3}& - \\ \hline
{AlexNet with channel-out \cite{alexchannel}}& {8} & {181.3} & - &{\textbf{13.2}} & -\\ \hline
{GoogLeNet \cite{cocotextpatch}} & {22} &  {41.2} & {9.4 }  & - & {9.9} \\ \hline
{AlexNet \cite{alexnet}, CaffeNet \cite{cocotextpatch}} & {8} &  {181.3} & {5} & {18.0} &{9.1} \\ \hline
\textbf{FFNet (the proposed model)}& {20} &  {\textbf{10.8}} & \textbf{2.8}& 13.6   &  {\textbf{9.0}} \\ \hline
\multicolumn{6}{l}{* Average forward path time per image on a K80 GPU} \\
\multicolumn{6}{l}{** CTP: COCO-Text-Patch dataset \cite{cocotextpatch}}

\end{tabular}
\end{table}

\section{Conclusion}
\label{section:conclusion}
This paper has presented a new concept, called \textit{input fast-forwarding}, which results in improved performance for deep-learning systems.
The approach utilizes parallel data paths that provide two advantages over previous approaches. 
One advantage is the explicit merging of
higher-level representations of data with lower-level representations.
A second advantage is a substantial reduction to the effects of the vanishing gradients problem.

To evaluate the model, we built a 20-layer network (FFNet) that implements the fast-forwarding concept. The network consists of simple convolutional layers, with no added complexities, to prove that the outstanding performance of the model is primarily the result of the fast-forwarding approach.
Empirical results also showed convergence during training at virtually the same rate as the bigger and more complex models.
FFNet achieved an error rate of 13.6\% on the CIFAR-10 dataset, which is on par with one variation of AlexNet.
When tested on COCO-Text-Patch, FFNet's performance surpassed that of CaffeNet and GoogLeNet, which are all significantly larger in size.

These results suggest that similar advantages may be obtained through the application of fast-forwarding to other models, and with different benchmark datasets.

\bibliographystyle{splncs}
\bibliography{isvc_submission}

\end{document}